\documentclass[runningheads]{llncs}
\usepackage{graphicx}
\usepackage[english]{babel}
\usepackage{tikz}
\usepackage{comment}
\usepackage{amsmath,amssymb} 

\usepackage{color}
\usepackage{multirow}
\usepackage[ruled,vlined]{algorithm2e}
\usepackage{algorithmic}
\usepackage[hidelinks,colorlinks=true]{hyperref}

\begin{document}
\newcommand{\keypoint}[1]{\noindent\textbf{#1}\quad}
\newcommand{\superkeypoint}[1]{\noindent\textbf{#1}\quad}
\newcommand{\ie}{\textit{i}.\textit{e}.}
\newcommand{\eg}{\textit{e}.\textit{g}.~}
\newcommand{\vs}{\textit{v}.\textit{s}.~}
\newcommand{\etc}{\textit{etc}}
\newcommand{\newadd}[1]{\textcolor{blue}{#1}}

\pagestyle{headings}
\mainmatter
\def\ECCVSubNumber{1986}  

\title{
Feature Representation Learning for Unsupervised Cross-domain Image Retrieval} 

\titlerunning{Feat. Rep. Learning for Unsupervised Cross-domain Image Retrieval}
%
\author{Conghui Hu\orcidID{0000-0002-4984-3960} \and
Gim Hee Lee\orcidID{0000-0002-1583-0475}}
\authorrunning{Conghui Hu and Gim Hee Lee}
%
\institute{Department of Computer Science, National University of Singapore\\
\email{\{conghui,gimhee.lee\}@nus.edu.sg}}
\maketitle

\begin{abstract}
Current supervised cross-domain image retrieval methods can achieve excellent performance. However, the cost of data collection and labeling imposes an intractable barrier to practical deployment in real applications. In this paper, we investigate the unsupervised cross-domain image retrieval task, where class labels and pairing annotations are no longer a prerequisite for training. This is an extremely challenging task because there is no supervision for both in-domain feature representation learning and cross-domain alignment. We address both challenges by introducing: 1) a new cluster-wise contrastive learning mechanism to help extract class semantic-aware features, and 2) a novel distance-of-distance loss to effectively measure and minimize the domain discrepancy without any external supervision. Experiments on the Office-Home and DomainNet datasets consistently show the superior image retrieval accuracies of our framework over state-of-the-art approaches. Our source code can be found at \url{https://github.com/conghuihu/UCDIR}.
 
\end{abstract}

\section{Introduction}
Cross-domain image retrieval refers to the task where the imagery data in one domain is used as query to retrieve the relevant samples in other domains. This task has many useful applications in our daily life. For example, sketch-based photo retrieval can be used in online shopping to search a product. To facilitate effective cross-domain retrieval, existing works takes the annotated class labels \cite{sangkloy2016sketchy} or cross-domain pairing information \cite{yu2016sketch} as supervision to train the model. However, it is always expensive and tedious to annotate labels for both domains, which severely limits the practical value of previous fully-supervised works. This limitation motivates us to circumvent the requirement for large amounts of annotated ground truth data by investigating the task of unsupervised cross-domain image retrieval. We thus focus on the category-level unsupervised cross-domain image retrieval. Specifically, the goal is to train a network to retrieve images from the same category using the query image across a different domain without any annotated class labels and cross-domain pairing information for the training data. Two challenges must be solved to achieve the goal of category-level unsupervised cross-domain image retrieval: 1) effectively bridge the gap between an imagery input and its corresponding semantic concept without label supervision, and 2) align the data between different domains without any cross-domain pair annotation.

\begin{figure}[!t]
\centering
\includegraphics[width=1\columnwidth]{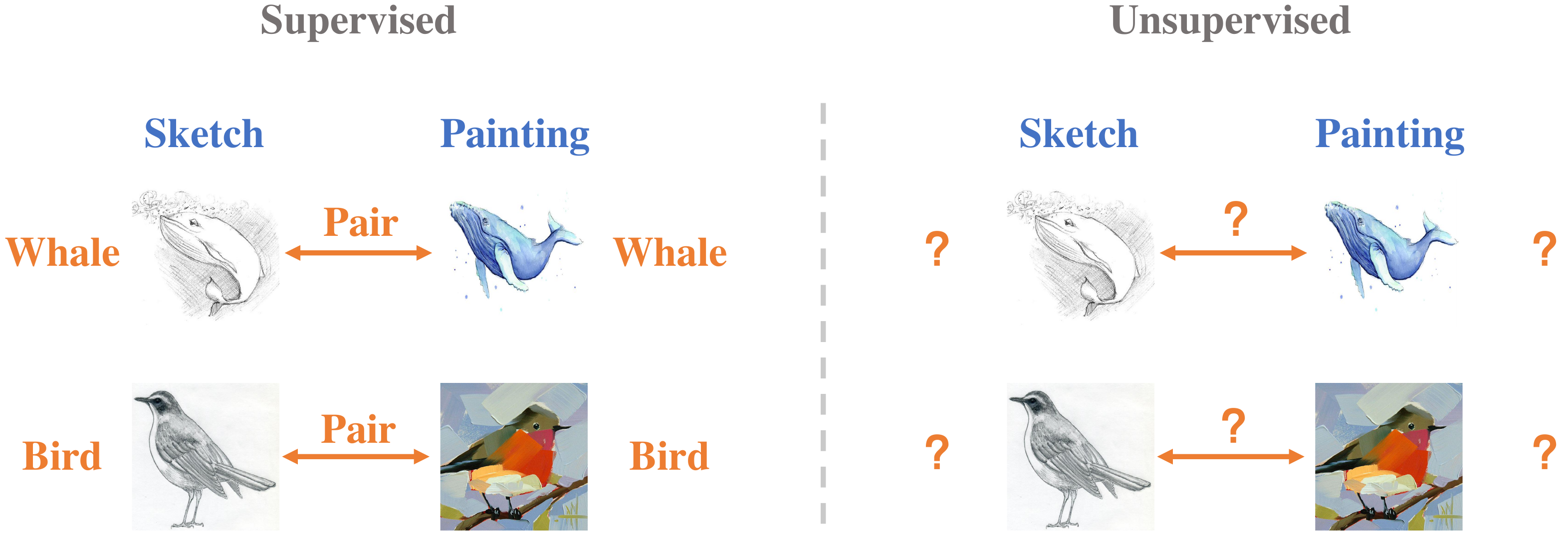}
\caption{Illustration of unsupervised cross-domain image retrieval. Compared with its supervised counterpart on the left, class label and pair annotation are not accessible in our unsupervised setting.}
\label{dd_illustration}
\end{figure}

The first challenge we faced is common in unsupervised feature representation learning \cite{wu2018unsupervised,caron2018deep,li2020prototypical}, whose objective is to extract discriminative feature representation from pixel-level input without class annotations. Nonetheless, unsupervised feature representation learning does not consider domain shifts and thus would fail catastrophically when directly applied to our category-level unsupervised cross-domain image retrieval task due to the domain gap.
The second challenge we faced is closely related to the line of works in unsupervised domain adaptation \cite{yue2021prototypical}, where an unlabeled target domain needs to be aligned with the source domain to accurately classify the data in the target domain. However, fully \cite{damodaran2018deepjdot} or partially \cite{yue2021prototypical} labeled source domain data are normally available for unsupervised domain adaptation. As a result, the task of unsupervised domain adaptation is easier than our category-level unsupervised cross-domain image retrieval since the labels in the source domain data can be used to learn discriminative features in the source domain and then transferred to the unlabeled target domain. Furthermore, our goal is to retrieve image of same category from the other domain, while unsupervised domain adaptation algorithms mostly focus on image classification. 

In this paper, we formulate a novel end-to-end learning framework which incorporates both in-domain unsupervised representation learning and cross-domain alignment
to accomplish our objective of training cross-domain image retrieval model in an unsupervised manner. We address the first challenge by introducing a new cluster-wise contrastive learning mechanism to help extract class semantic-aware features. In contrast to existing instance-wise contrastive learning loss \cite{oord2018representation} which neglects class semantic by only considering the augmented views of itself as positive samples, our cluster-wise contrastive loss is based on feature clusters that pulls samples of similar semantics closer and pushes different clusters apart.
We address the second challenge by proposing a novel distance-of-distance loss which is able to effectively measure and minimize the domain discrepancy without external supervision. Specifically, we circumvent the difficulty of domain alignment due to the unknown class labels associated to the feature clusters in each domain by designing our distance-of-distance loss to be invariant to the cluster orders.   

Our main contributions can be summarized as the follows:
\begin{enumerate}
    \item We develop a novel feature representation learning algorithm for unsupervised cross-domain image retrieval. 
    \item To enable semantic-aware feature extraction, we propose a new cluster-wise contrastive learning loss to minimize the feature distances between semantically similar samples.
    \item A novel distance-of-distance loss is carefully designed to measure domain discrepancy and help achieve cross-domain alignment.
    \item Extensive experiments on the Office-Home and DomainNet datasets demonstrate the efficacy of our proposed framework.
\end{enumerate}

\section{Related Work}
\subsection{Cross-domain Image Retrieval}
Image retrieval is a long-standing research problem in computer vision. Given a query image, the objective is to retrieve images that meet certain predefined criteria from the database \cite{smeulders2000content}. The criteria can be category-level or instance-level correspondence according to the granularity. All images in the database are ranked according to the distance to the query image \cite{muller2001performance}. Thus, the task boils down to effectively measure the similarity between images. It becomes more challenging when the query and images in database are from different domains, \ie, cross-domain image retrieval. For example, the query can be just a free-hand sketch with a few strokes, while images in the database are all real photographs \cite{sangkloy2016sketchy}. To accurately compute the distance for images across domains, class labels are required as the supervision \cite{sangkloy2016sketchy} for class semantic-aware feature extraction and cross-domain pairing annotations are leveraged to bridge the domain gap by minimizing triplet \cite{yu2016sketch,zhao2021m3l} or HOLEF loss \cite{song2017deep}. However, the annotations used as supervision for model training are always labour-intensive to source. We are therefore motivated to investigate the unsupervised cross-domain image retrieval that shares all challenges laid out in cross-domain image retrieval but without external supervision. 

\subsection{Unsupervised Representation Learning}
Unsupervised representation learning has been actively studied in recent years. Deep clustering related methods attempt to assign a pseudo class label for each sample by traditional clustering methods such as K-means \cite{caron2018deep} or maintain a set of trainable cluster centroids \cite{gao2020deep}, and pseudo labels are continually refined during training. In contrast to deep clustering, instance discrimination \cite{wu2018unsupervised} directly regards every sample itself as a standalone class and the training objective is to distinguish the sample from all the rest of data for the sake of extracting meaningful discriminative features. Self-supervised learning approaches facilitate the representation learning by introducing various pretext tasks like image rotation prediction \cite{gidaris2018unsupervised}, jigsaw puzzle solving \cite{noroozi2016unsupervised} and image in-painting \cite{pathak2016context}. Supervisions for all these pretext tasks are free to obtain. Contrastive learning has been increasingly gaining popularity in unsupervised representation learning due to its effectiveness. The goal of contrastive learning algorithms is to maximize the agreement between positive pairs like two augmented views of the same images \cite{chen2020simple} or the image and its corresponding cluster centroid \cite{li2020prototypical}. Nevertheless, the aforementioned unsupervised representation learning methods are originally devised for single domain data. The large gap between different domains preemptively limits their practical value.

\subsection{Unsupervised Domain Adaptation}
A related line of research that addresses the domain gap without full-supervision is unsupervised domain adaptation. Conventional unsupervised domain adaptation targets on transferring knowledge learned from an annotated source domain to a novel unlabeled target domain. The model can be trained to predict semantic-aware features for the source-domain data with the help of source domain label. The key challenge then becomes aligning the target domain data with their counterpart in the source domain. The domain discrepancy can be directly measured by Maximum Mean Discrepancy (MMD) \cite{gretton2012kernel} or Joint MMD \cite{long2017deep}, and minimized to remedy the domain gap \cite{long2016unsupervised,long2017deep}. \cite{shen2018wasserstein,damodaran2018deepjdot} managed to search for matching data pairs across domains through Optimal Transport \cite{villani2009optimal}. Moreover, Generative adversarial Networks \cite{goodfellow2014generative} can also be utilized to break the domain barrier in either feature-level \cite{ganin2016domain} or pixel-level \cite{hoffman2018cycada}. More recently, a new domain adaptation setting with only few-shot annotated samples from source domain is introduced to
further reduce the data labeling cost. \cite{kim2020cross,kim2021cds} employed instance discrimination \cite{wu2018unsupervised} for both in-domain and cross-domain to learn a shared and instance-discriminative feature space. In \cite{yue2021prototypical}, prototypes are applied to make the feature more semantic-aware. Our unsupervised cross-domain image retrieval is different from domain adaptation in two aspects: 1) There is no requirement for labeled data. Both source and target domain are unlabeled; 2) Our task is image retrieval rather than image classification.

\section{Our Method}
\subsection{Overview}
We use domain $A$ and domain $B$ to denote the domain shift in images.
Given a query image $I_i^A$ of category $k$ in domain $A$, category-level cross-domain image retrieval is considered successful when images of the same category $k$ in domain $B$ are retrieved. To accomplish the goal of cross-domain retrieval, it is required to train a valid feature extractor $f_\theta : I \mapsto \mathbf{x}$ which can project input image $I$ from both domains to feature $\mathbf{x}$ in a common embedding space. All images $\mathcal{I}^B = {\left\{I_j^B\right\}}_{j=1}^M$ in domain $B$ are then ranked by the feature distance $d(\mathbf{x}_i^A, \mathbf{x}_j^B)$ between $I_j^B$ and the query image $I_i^A$ in domain $A$. For $I_i^B$ of category $k$ to appear on top of the list, feature extractor $f_\theta$ needs to be capable of learning: 1) class semantic-aware features to discriminate samples among different classes; and 2) domain-agnostic features to facilitate the direct distance measurement between images in domain $A$ and $B$. 
However, only a set of unlabeled images $\mathcal{I}^A = {\left\{I_i^A\right\}}_{i=1}^N$ and $\mathcal{I}^B = {\left\{I_j^B\right\}}_{j=1}^M$ from domain $A$ and $B$ are provided for training under the unsupervised cross-domain image retrieval setting. As a result, learning semantically meaningful and domain-invariant feature embeddings becomes extremely challenging. To learn feature representations that fulfill the aforementioned requirements, we design our framework according to the following two aspects: 1) In-domain representation learning which targets on learning class-discriminative features through a novel cluster-wise contrastive learning mechanism; 2) Cross-domain alignment through the distance of distance minimization as shown in Figure \ref{dd_illustration}.

\subsection{In-domain Cluster-wise Contrastive learning}
Instance-wise contrastive learning methods take only augmented views of the same instance as positive pair, while all other samples in the dataset are regarded as the negatives. The loss function \cite{oord2018representation} is defined as:
\begin{equation}
\label{p-eq}
\mathcal{L}_\text{IW} = \sum\limits_{i \in \mathbf{I}} -\log \frac{\exp(\mathbf{x}^{\top}_{i}\mathbf{x}'_{i}/{\tau})}{\sum\limits_{a \in \mathbf{I}}\exp(\mathbf{x}^{\top}_{i}\mathbf{x}'_{a}/{\tau})},
\end{equation}
where $\mathbf{x}_{i}$ and $\mathbf{x}'_{i}$ can be feature embeddings from different augmented views of instance $i$. $\mathbf{I}$ represents indices of all the samples in the same domain.

Although all instances are well-seperated at instance-level after training the feature extractor with $\mathcal{L}_\text{IW}$, they are not clustered together according to their classes. However, the feature space that can encode the class semantic structure of the data is desired in the cross-domain image retrieval task. This motivates us to design the cluster-wise contrastive learning loss. Specifically, there are two steps in our cluster-wise contrastive learning: 1) image clustering and 2) contrastive learning with pseudo label. Additionally, we propose to perform cluster-wise contrastive learning separately in domain $A$ and $B$. For brevity, we remove the domain notation in this section.

\vspace{1mm}
\keypoint{Image clustering.} Following MoCo \cite{chen2020mocov2}, we maintain a momentum encoder $f_{{\theta}^{'}}$ to extract features for image clustering as it yields more consistent clusters. Let $\mathbf{x}_{i}$ and $\mathbf{x}'_{i}$ be the features extracted from the trainable encoder $f_{\theta}$ and momentum encoder $f_{{\theta}'}$, respectively. We apply $K$-means on all image features $\left\{\mathbf{x}'_{i}\right\}_{i=1}^N$ in one single domain to obtain its $K$ clusters. Each sample $I_i$ is assigned with a pseudo label $y_i$ according to the $K$-means results. All pseudo labels are updated after every epoch.

\vspace{1mm}
\keypoint{Contrastive learning with pseudo label.} Given the pseudo labels predicted by image clustering, we can now conduct cluster-wise contrastive learning. Samples in the same cluster as the query are used to form positive pairs. As a result, the feature extractor is trained to pull feature distances within the same cluster closer, while pushing different clusters apart. Specifically, the learning objective of cluster-wise contrastive learning is given by:
\begin{equation}
\label{p-eq}
\mathcal{L}_\text{CW} = \sum\limits_{i \in \mathbf{I}} \frac{-1}{{|\mathbf{P}(i)|}} \sum\limits_{p \in \mathbf{P}(i)} \log \frac{\exp(\mathbf{x}^{\top}_{i}\mathbf{x'}_{p}/{\tau})}{\sum\limits_{a \in \mathbf{I}}\exp(\mathbf{x}^{\top}_{i}\mathbf{x'}_{a}/{\tau})},
\end{equation}
where $\mathbf{I}$ denotes indices of all the samples in the same domain and $\mathbf{P}(i)$ represents the indices for the set of samples belonging to the same cluster as $I_i$, \ie, $\mathbf{P}(i) = \left\{p \in \mathbf{I}: y_p = y_i \right\}$ and $|\mathbf{P}(i)|$ is its cardinality.

To encourage local smoothness and generate valid clustering results at the beginning of the training process, we also add the instance-wise contrastive loss. Therefore, our loss for in-domain feature representation learning becomes:
\begin{equation}
\mathcal{L}_\text{in-domain} = \mathcal{L}_\text{IW} + \lambda\mathcal{L}_\text{CW}.
\end{equation}
Since the clustering results are not sufficient reliable at the initial learning stage, we gradually increase the weight for $\mathcal{L}_{CW}$ and set $\lambda$ as:

\begin{equation}
\lambda = \left\{
\begin{array}{lll}
0 & \quad ep <= T_1 \\
\alpha\frac{ep - T_1}{T_2 - T_1}&\quad T_1 < ep < T_2\\
\alpha & \quad ep >= T_2
\end{array}
\right.,
\end{equation}
where $\alpha$ is a weight hyper-parameter. $ep$, $T_1$ and $T_2$ are the current training epoch, the number of epoch to include $\mathcal{L}_{CW}$, the number of epoch to stop increasing the weight for $\mathcal{L}_{CW}$, respectively.

\subsection{Cross-domain Alignment}
Domain-invariant is another requirement for the features in cross-domain image retrieval. However, it is difficult to effectively align feature clusters across domains when there is no class label nor correspondence annotation that can be utilized as supervision in our unsupervised setting. 
Furthermore, the order of the predicted cluster centroids 
$\left\{\mathbf{c}^A_{u}\right\}_{u=1}^{K}$ and $\left\{\mathbf{c}^B_{u}\right\}_{u=1}^{K}$ for domain $A$ and $B$ are not fixed since we perform $K$-means separately in the two domains. The unknown correspondences between the clusters across the two domains increases the challenge for cross-domain alignment. To solve this problem, we propose to measure the cross-domain distance of the in-domain distance. In other words, the Distance-of-Distance (DD) loss.

\vspace{1mm}
\keypoint{In-domain Distance.} Given an input image $I_i$, we calculate its clustering probabilities using the cluster centroids, \ie:
\begin{equation}
p_{i}^{u} = \frac{\exp(\mathbf{x}^{\top}_{i}\mathbf{c}_{u}/{\phi})}{\sum\limits_{k=1}^{K}\exp(\mathbf{x}^{\top}_{i}\mathbf{c}_{k}/{\phi})},
\end{equation}
where $\phi$ is a temperature hyper-parameter. $\mathbf{c}_{u}$ represents the centroids for cluster $u$. The clustering probabilities for $I_i$ is $\mathbf{p}_{i} = {[p_{i}^{({1})}, p_{i}^{({2})}..., p_{i}^{({K})}]}^{\top}$. We leverage the centroids for domain $A$ and $B$ to obtain $\mathbf{p}_{i}^{A}$ and $\mathbf{p}_{i}^{B}$, respectively. The in-domain distance for domain $A$ is defined as:
\begin{equation}
\label{d-eq}
\begin{split}
d_{{ij}}^A &= \operatorname{D}(\mathbf{p}_{i}^A, \mathbf{p}_{j}^A), \quad \text{where} \\ 
\mathbf{p}_{i}^A &= {[p_{i}^{(A^{1})}, p_{i}^{(A^{2})}..., p_{i}^{(A^{K})}]}^{\top}, \quad
\mathbf{p}_{j}^A = {[p_{j}^{(A^{1})}, p_{j}^{(A^{2})}..., p_{j}^{(A^{K})}]}^{\top}.
\end{split}
\end{equation}
Here, $\operatorname{D}(\cdot,\cdot)$ is the cosine distance. This in-domain distance measures the difference in clustering probabilities for samples. 
\begin{proposition}
Value of $d_{{ij}}^A$ \textit{remains the same} when the order of the elements in $\mathbf{p}_{i}^A$ and $\mathbf{p}_{j}^A$ are simultaneously shuffled, \ie, order of the centroids is changed.
\end{proposition}
\begin{proof}
Suppose we randomly shuffle corresponding elements of $\mathbf{p}_{i}^A$ and $\mathbf{p}_{j}^A$ to:
\begin{equation*}
\mathbf{p'}_{i}^A = {[p_{i}^{(A^{K})}, p_{i}^{(A^{1})}, \cdots, p_{i}^{(A^{2})}]}^{\top}~~\text{and}~~\mathbf{p'}_{j}^A = {[p_{j}^{(A^{K})}, p_{j}^{(A^{1})}, \cdots, p_{j}^{(A^{2})}]}^{\top},
\end{equation*}
we get ${d'}_{ij}^A = \operatorname{D}(\mathbf{p'}_{i}^A, \mathbf{p'}_{j}^A) = 1 - {\eta'}^{-1}(p_{i}^{(A^{K})}p_{j}^{(A^{K})} + p_{i}^{(A^{1})}p_{j}^{(A^{1})} + \cdots + p_{i}^{(A^{2})}p_{j}^{(A^{2})}) = 1 - \eta^{-1}(p_{i}^{(A^{1})}p_{j}^{(A^{1})} + p_{i}^{(A^{2})}p_{j}^{(A^{2})} + \cdots + p_{i}^{(A^{K})}p_{j}^{(A^{K})}) = \operatorname{D}(\mathbf{p}_{i}^A, \mathbf{p}_{j}^A) = {d}_{ij}^A$, where $\eta' = \| \mathbf{p'}_{i}^A\| \|\mathbf{p'}_{j}^A\| = \| \mathbf{p}_{i}^A\| \|\mathbf{p}_{j}^A\| = \eta$ due to the commutative property of addition. 
\qed
\end{proof}
\begin{corollary}
Our proposed in-domain distance has the important property of \textit{order-invariant}. This order-invariant property also holds for ${d}_{{ij}}^B$ for domain $B$.
\end{corollary}

\begin{figure}[!t]
\centering
\includegraphics[width=1\columnwidth]{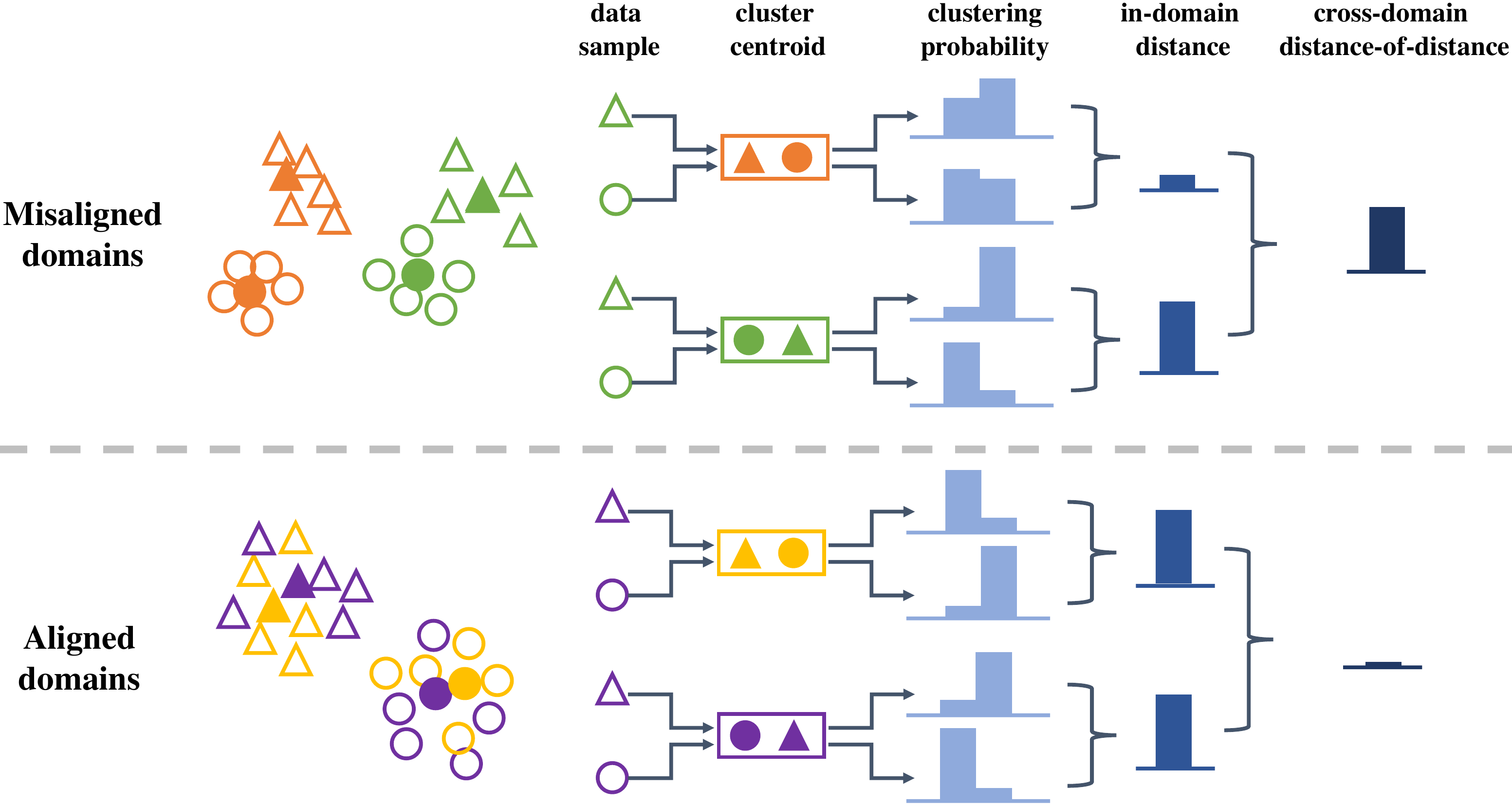}
\caption{Illustration of cross-domain distance-of-distance loss. Shapes and colors represent the data samples of different classes and domains, respectively.}
\label{dd_illustration}
\end{figure}

\keypoint{Cross-domain Distance-of-Distance.} As mentioned previously, our key challenge is to design a proper discrepancy measurement method to align the two domains while the cluster orders are unknown. Provided with the order-invariant in-domain distance, we devise a new cross-domain distance of distance:
\begin{equation}
dd_{{ij}} = \operatorname{DD}({d}_{{ij}}^A, d_{{ij}}^B).
\end{equation}
where $\operatorname{DD}(\cdot,\cdot)$ represents the distance-of-distance (DD) calculator that measures the L2 distance between two in-domain distances. As ${d}_{{ij}}^A$ and $d_{{ij}}^B$ are both order-invariant and the data samples used in the in-domain distance calculation are the same $I_i$ and $I_j$. The value of $dd_{{ij}}$ is then only related to the difference between values in the two sets of centroids $\left\{\mathbf{c}^A_{u}\right\}_{u=1}^{K}$ and $\left\{\mathbf{c}^B_{u}\right\}_{u=1}^{K}$ regardless of the order of cluster centroids. 
For two well-aligned domains $A$ and $B$, $dd_{{ij}}$ is small since the centroids for the two domains are similar. However, the corresponding centroids becomes disparate when there is a big difference between the feature distribution for domain $A$ and $B$. Consequently,
the value of ${d}_{{ij}}^A$ and ${d}_{{ij}}^B$ would differ greatly and thus lead to a large $dd_{{ij}}$. The detailed illustration can be found in Figure \ref{dd_illustration}. We thus propose our $\operatorname{DD}$ loss to effectively measure the discrepancy between features in two domains, where a smaller DD loss indicates better alignment. Formally, our $\operatorname{DD}$ loss is written as:
\begin{equation}
\mathcal{L}_\text{DD} = \sum\limits_{i \in \mathbf{R}^A}\sum\limits_{j \in \mathbf{R}^A}dd_{{ij}} + \sum\limits_{i \in \mathbf{R}^B}\sum\limits_{j \in \mathbf{R}^B}dd_{{ij}},
\end{equation}
where $\mathbf{R}^A$ and $\mathbf{R}^B$ contains indices for domain $A$ and domain $B$ data in current batch. $i$ and $j$ are indices for instances.

\vspace{1mm}
\keypoint{Entropy minimization.} Our $\operatorname{DD}$ loss can also be trivially minimized when all the clustering probabilities are uniformly distributed, \ie, all the elements the in $\mathbf{p}_{i}^A$ and $\mathbf{p}_{i}^B$ are the same. To prevent this trivial solution, we propose to minimize the self-entropy for all clustering probabilities:

\begin{equation}
\mathcal{L}_\text{SE} = \sum\limits_{i \in \mathbf{I}^A} (H(\mathbf{p}_{i}^A) + H(\mathbf{p}_{i}^B)) + \sum\limits_{j \in \mathbf{I}^B} (H(\mathbf{p}_{j}^A) + H(\mathbf{p}_{j}^B)),
\end{equation}
where $\mathbf{I}^A$ and $\mathbf{I}^B$ are the indices for all samples in domain $A$ and $B$, respectively. Consequently, our training objective for cross-domain alignment is given by:
\begin{equation}
    \mathcal{L}_\text{cross-domain} = \beta\mathcal{L}_\text{DD} + \gamma\mathcal{L}_\text{SE},
\end{equation}
where $\beta$ and $\gamma$ are hyper-parameters to balance the two loss terms. 

\subsection{Summary}
To facilitate the feature extractor $f_\theta$ training for cross-domain image retrieval without any labeled data as supervision, we introduce a new cluster-wise contrastive learning loss for semantic-aware feature extraction, and propose a novel $\operatorname{DD}$ loss to effectively evaluate whether the two domains are aligned and train the feature extractor $f_\theta$ to minimize the $\operatorname{DD}$ loss. Our final training target is defined as:
\begin{equation}
    \mathcal{L}_\text{all} = \mathcal{L}_\text{in-domain} + \mathcal{L}_\text{cross-domain}.
\end{equation}

\section{Experiments}
\subsection{Datasets and Settings}
\keypoint{Datasets.} Our proposed method is comprehensively evaluated on two datasets: 1) Office-Home \cite{venkateswara2017deep} offers 4 domains (Art, Clipart, Product, Real) with 65 categories. 2) DomainNet \cite{peng2019moment} with six different domains (Clipart, Infograph, Painting, Quickdraw, Real and Sketch). We use all six domains and select those categories with more than 200 images in every domain for training and testing. According to this criterion, 7 categories are used in our experiments.
\begin{table}[!t]
\begin{center}
\caption{Unsupervised Cross-domain Retrieval Accuracy $(\%)$ on Office-Home.}
\scalebox{1}{
\begin{tabular}{r|c|c|c|c|c|c|c|c|c}
\hline
\multirow{2}{*}{Method}

\multirow{2}{*}{}
& \multicolumn{3}{c|}{Art$\rightarrow$Real} & \multicolumn{3}{c|}{Real$\rightarrow$Art}& \multicolumn{3}{c}{Art$\rightarrow$Product}\\
\cline{2-10}
& P@1 & P@5 & P@15 & P@1 & P@5 & P@15 & P@1 & P@5 & P@15 \\
\hline
ID \cite{wu2018unsupervised} & 35.89 & 33.13 & 29.60 & 39.89 & 34.42 & 27.65 & 25.88 & 24.91 & 22.49 \\
ProtoNCE \cite{li2020prototypical} & 40.50 & 36.39 & 34.00  & 44.53 & 39.26 & 32.99 & 29.54& 27.89 & 25.75 \\
CDS \cite{kim2021cds} & 45.08 & 41.15 & 38.73 & 44.71 & 40.75 & 35.53 & 32.76 & 31.47 & 28.90 \\
PCS \cite{yue2021prototypical} & 41.70 & 38.51  & 36.22 & 44.96 & 39.88 & 33.99 & 33.29 & 31.50 & 29.53 \\
Ours  & \textbf{45.12} & \textbf{42.33} & \textbf{40.06}& \textbf{47.95} & \textbf{43.68} & \textbf{38.38} & \textbf{35.39} & \textbf{34.67} & \textbf{32.61}\\
\hline

\multirow{2}{*}{}
& \multicolumn{3}{c|}{Product$\rightarrow$Art} & \multicolumn{3}{c|}{Clipart$\rightarrow$Real}& \multicolumn{3}{c}{Real$\rightarrow$Clipart}\\
\cline{2-10}
& P@1 & P@5 & P@15 & P@1 & P@5 & P@15 & P@1 & P@5 & P@15 \\
\hline
ID \cite{wu2018unsupervised} & 32.17 & 25.94 & 20.23 & 29.48 & 26.48 & 23.25 & 35.51 & 32.17 & 27.96 \\
ProtoNCE \cite{li2020prototypical} & 35.73 & 30.61 &24.55  & 25.25 & 22.66 &20.83 & 41.15 & 37.66 & 31.95\\
CDS \cite{kim2021cds} & 35.75 & 32.48 & 26.82 & 32.51 & 30.30 & 27.80 & 38.88 & 36.48 & 33.16  \\
PCS \cite{yue2021prototypical} & 39.24 & 34.77 & 28.77  & 29.07 & 26.06 & 24.00 & 40.60 & 38.11 & 34.06  \\
Ours & \textbf{42.51} & \textbf{37.94} & \textbf{31.41}& \textbf{33.31} & \textbf{30.57} & \textbf{28.14}& \textbf{44.66} & \textbf{41.47} & \textbf{37.41}  \\
\hline

& \multicolumn{3}{c|}{Product$\rightarrow$Real} & \multicolumn{3}{c|}{Real$\rightarrow$Product}& \multicolumn{3}{c}{Product$\rightarrow$Clipart}\\
\cline{2-10}
& P@1 & P@5 & P@15 & P@1 & P@5 & P@15 & P@1 & P@5 & P@15 \\
\hline
ID \cite{wu2018unsupervised} & 50.73 & 45.03 & 39.05 & 45.12 & 41.46 & 38.01 & 31.52 & 28.55 & 24.15 \\
ProtoNCE \cite{li2020prototypical} & 53.84 & 48.25 & 42.21  & 47.74 & 44.85 & 41.21 & 36.13& 33.99 & 28.24\\
CDS \cite{kim2021cds} & 54.00 & 50.07 & 45.60 & 49.39 & 47.27 & 43.98 & 37.69 & 34.99 & 30.42 \\
PCS \cite{yue2021prototypical} & 56.45 & 50.78 & 45.37 & 49.90 & 47.11 & 43.73 & 39.51 & \textbf{37.51} & 32.81 \\
Ours  & \textbf{57.42} & \textbf{52.69} & \textbf{47.90} & \textbf{51.71} & \textbf{48.48}& \textbf{44.95}& \textbf{42.26} & 37.42 & \textbf{33.74} \\
\hline
\multirow{2}{*}{}
& \multicolumn{3}{c|}{Clipart$\rightarrow$Product} & \multicolumn{3}{c|}{Art$\rightarrow$Clipart}& \multicolumn{3}{c}{Clipart$\rightarrow$Art} \\
\cline{2-10}
& P@1 & P@5 & P@15 & P@1 & P@5 & P@15 & P@1 & P@5 & P@15 \\
\hline
ID \cite{wu2018unsupervised} & 24.01 & 22.42 & 20.60 & 26.78 & 24.79 & 21.64 & 21.17 & 17.86 & 14.71 \\
ProtoNCE \cite{li2020prototypical} & 21.17 & 20.63 & 20.47  & 28.97 & 26.15 & 22.98 & 21.33 & 17.40 & 14.46 \\
CDS \cite{kim2021cds} & 27.24 & 26.46 & 24.86 & 25.59 & 23.77 & 22.41 & 22.41 & 20.34 & 17.34 \\
PCS \cite{yue2021prototypical} & 26.39 & 25.86 & 24.92 & 31.23 & 28.74 & 26.11 & 24.51 & 21.27 & 17.54\\
Ours & \textbf{27.79} & \textbf{27.26} & \textbf{25.97} & \textbf{32.67} & \textbf{30.79} & \textbf{28.70}& \textbf{27.26} & \textbf{23.94} & \textbf{20.53} \\
\hline

\multirow{6}{*}{Average}
& \multicolumn{3}{c|}{ID \cite{wu2018unsupervised}} & \multicolumn{3}{c|}{ProtoNCE \cite{li2020prototypical}}& \multicolumn{3}{c}{CDS \cite{kim2021cds}} \\
\cline{2-10}
& P@1 & P@5 & P@15 & P@1 & P@5 & P@15 & P@1 & P@5 & P@15 \\
\cline{2-10}
& 33.18 & 29.76 & 25.78 & 35.49 & 32.15 & 28.30 & 37.17 & 34.63 & 31.30 \\
\cline{2-10}
 & \multicolumn{3}{c|}{PCS \cite{yue2021prototypical}} & \multicolumn{3}{c|}{Ours}& \multicolumn{3}{c}{Improvement} \\
\cline{2-10}
& P@1 & P@5 & P@15 & P@1 & P@5 & P@15 & P@1 & P@5 & P@15 \\
\cline{2-10}
& 38.07 & 35.01 & 31.42 & \textbf{40.67} & \textbf{37.60} & \textbf{34.15} & \textbf{+2.60} & \textbf{+2.59} & \textbf{+2.73}\\
\hline
\end{tabular}
}
\label{office-retrieval}
\end{center}
\end{table}

\keypoint{Implementation details.}
ResNet-50 \cite{he2016deep} is employed as the architecture for our feature extractor $f_\theta$. Both features $\mathbf{x}_i$ and cluster centroids $\mathbf{c}_u$ are L2 normalized 128-d vectors. To make sure the whole training procedure is fully unsupervised, we use parameters from the MoCov2 \cite{chen2020mocov2} model trained with unlabeled ImageNet dataset \cite{deng2009imagenet} to initialize $f_\theta$. The initial learning rate is 0.0002. We follow MoCov2 \cite{chen2020mocov2} to use the cosine learning rate schedule that gradually decreases the learning rate to 0. The total number of training epoch is 200. We adopt SGD to update the parameters in the feature extractor with momentum of 0.9 and a batch size of 64. Our framework is built with deep learning library Pytorch \cite{paszke2019pytorch}. $T_1$ and $T_2$ are set to 20 and 100 respectively. The cluster number $K$ is set to be the same as the number of classes in the training set, \ie, 65 for Office-Home and 7 for DomainNet.

\vspace{1mm}
\keypoint{Evaluation metrics.} To validate the retrieval performance for the Office-Home dataset, we follow \cite{kim2021cds} to calculate the precision among top 1/5/15 retrieved images as the minimum number of images for one single category is 15. As for DomainNet, we filter out those categories with fewer than 200 images. Thus, we measure the precision for top 50/100/200 retrieved image to provide a more comprehensive evaluation for retrieval accuracy in DomainNet. Since our task is category-level cross-domain retrieval, the retrieved images with the same semantic class as the query are regarded as the correct ones.

\vspace{1mm}
\keypoint{Baselines.} We use the following works as the baselines to evaluate our proposed method: 1) \textbf{ID} \cite{wu2018unsupervised} achieves unsupervised representation learning by instance discrimination where all instance are well-separated regardless of the category. 2) \textbf{ProtoNCE} \cite{li2020prototypical} is a more recent unsupervised representattion learning algorithm which proposes to use prototypes to help encode semantic structure of data and predict more aggregated feature clusters. 3) \textbf{CDS} \cite{kim2021cds} is originally designed for cross-domain self-supervised pre-training. In-domain instance discrimination and cross-domain instance matching are designed for learning a shared embedding space across domains. 4) \textbf{PCS} \cite{yue2021prototypical} is a cross-domain self-supervised learning approach that uses prototypical contrastive learning for in-domain feature learning and instance-prototype matching for cross-domain alignment. 

\begin{figure}[!t]
\centering
\includegraphics[width=1\columnwidth]{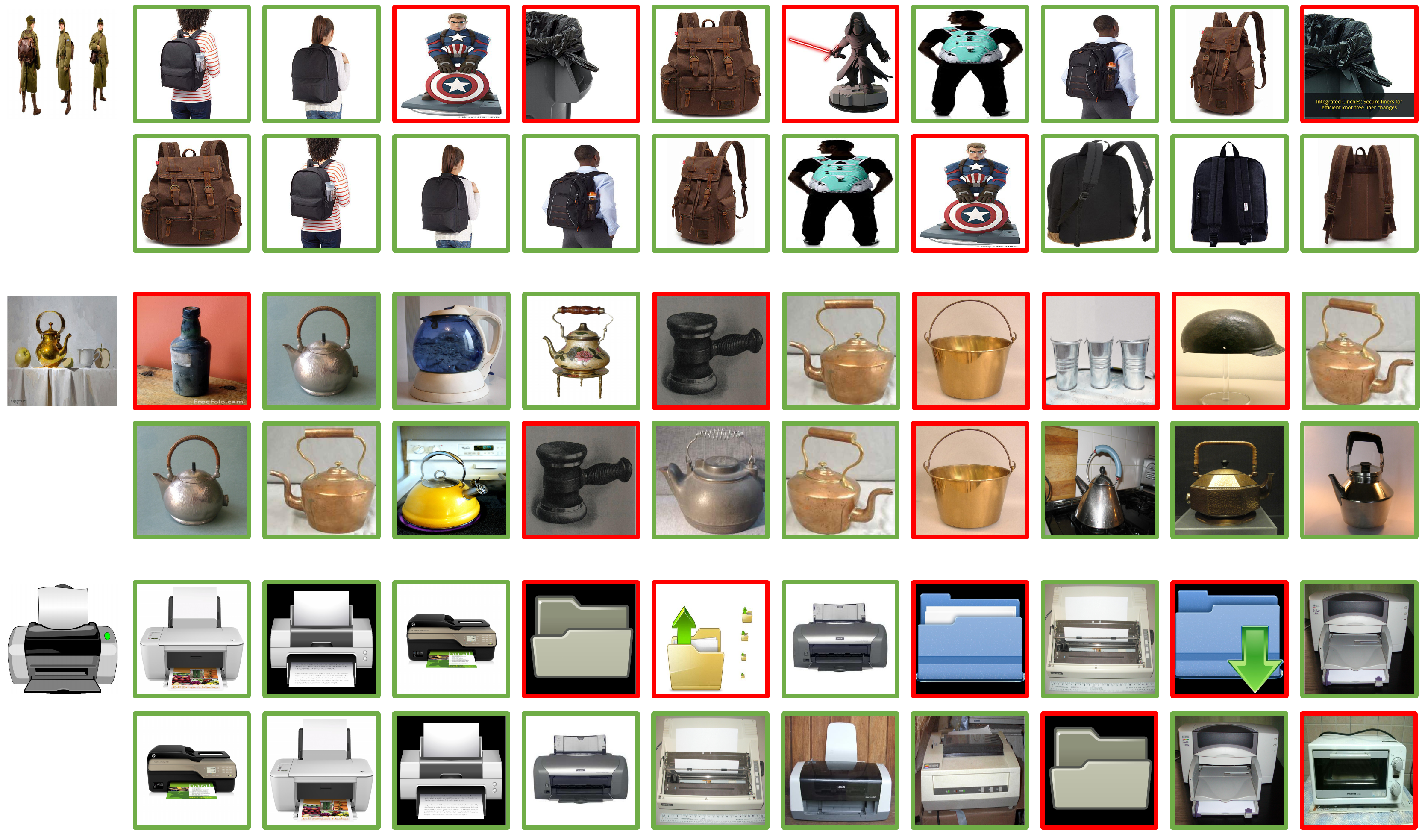}
\label{office-home-retrieval-vis}
\caption{Top 10 retrieval results on Office-Home. Row 1, 3, 5: Retrieval results of the best baseline method - PCS \cite{yue2021prototypical}; Row 2, 4, 6: Retrieval results of our framework. The green and red boxes indicate correct and incorrect retrievals, respectively.}
\end{figure}

\subsection{Results}
\superkeypoint{i) The Office-Home Dataset}

\vspace{1mm}
\keypoint{Settings.} Since there are four domains in the Office-Home dataset, we have altogether 6 different pairs (Art-Real, Art-Product, Clipart-Real, Product-Real, Product-Clipart, Art-Clipart) by matching any two of the four domains. Furthermore, the two domains in one pair can be both regarded as query domain to retrieve images from the other domain.

\begin{table}[!t]
\begin{center}
\caption{Unsupervised Cross-domain Retrieval Accuracy $(\%)$ on DomainNet.}
\scalebox{1}{
\begin{tabular}{r|c|c|c|c|c|c|c|c|c}
\hline
\multirow{2}{*}{Method}
& \multicolumn{3}{c|}{Clipart$\rightarrow$Sketch} & \multicolumn{3}{c|}{Sketch$\rightarrow$Clipart}& \multicolumn{3}{c}{Infograph$\rightarrow$Real}\\
\cline{2-10}
& P@50 & P@100 & P@200 & P@50 & P@100 & P@200& P@50 & P@100 & P@200 \\
\hline
ID \cite{wu2018unsupervised} & 49.46 & 46.09 & 40.44 & 54.38 & 47.12 & 37.73 & 28.27 & 27.44 & 26.33 \\
ProtoNCE \cite{li2020prototypical} & 46.85 & 42.67 & 36.35 & 54.52 & 45.04 & 35.06& 28.41& 28.53 & 28.50 \\
CDS \cite{kim2021cds} & 45.84 & 42.37 & 37.16 & 59.13 & 48.83 & 37.40 & 28.51 & 27.92 & 27.48 \\
PCS \cite{yue2021prototypical} & 51.01 & 46.87 & 40.19 & 59.70 & 50.67 & 39.38 & 30.56 & 30.27 & 29.68 \\
Ours  & \textbf{56.31} & \textbf{52.74} & \textbf{47.38} & \textbf{63.07} & \textbf{57.26}& \textbf{48.17}& \textbf{35.52} & \textbf{35.24} & \textbf{34.35} \\
\hline

\multirow{2}{*}{}
& \multicolumn{3}{c|}{Real$\rightarrow$Infograph} & \multicolumn{3}{c|}{Infograph$\rightarrow$Sketch}& \multicolumn{3}{c}{Sketch$\rightarrow$Infograph}\\
\cline{2-10}
& P@50 & P@100 & P@200 & P@50 & P@100 & P@200& P@50 & P@100 & P@200 \\
\hline
ID \cite{wu2018unsupervised} & 39.98 & 31.77 & 24.84 & 30.35 & 29.04 & 26.55 & 42.20 & 34.94 & 27.52 \\
ProtoNCE \cite{li2020prototypical} & 57.01 & 41.84 & 30.33 & 28.24 & 26.79 & 24.23 & 39.83 & 31.99 & 24.77 \\
CDS \cite{kim2021cds} & 56.69 & 39.76 & 26.38 & 30.55 & \textbf{29.51} & \textbf{27.00} & \textbf{46.27} & 36.11 & 27.33 \\
PCS \cite{yue2021prototypical} & 55.42 & 42.13 & 30.76  & 30.27 & 28.36 & 25.35 & 42.58 & 34.09 & 25.91\\
Ours & \textbf{57.74} & \textbf{46.69} & \textbf{35.47} & \textbf{31.29} & 29.33 & 26.54& 43.66 & \textbf{36.14} & \textbf{28.12} \\
\hline

\multirow{2}{*}{}
& \multicolumn{3}{c|}{Painting$\rightarrow$Clipart} & \multicolumn{3}{c|}{Clipart$\rightarrow$Painting}& \multicolumn{3}{c}{Painting$\rightarrow$Quickdraw}\\
\cline{2-10}
& P@50 & P@100 & P@200 & P@50 & P@100 & P@200& P@50 & P@100 & P@200 \\
\hline
ID \cite{wu2018unsupervised} & 64.67 & 54.41 & 40.07 & 42.37 & 39.61 & 35.56 & 20.34 & 19.59 & 18.79 \\
ProtoNCE \cite{li2020prototypical} & 55.44 & 43.74 &  32.59 & 39.13 & 35.87 & 32.07 & 21.63 & 21.24 & 20.56 \\
CDS \cite{kim2021cds} & 63.15 & 47.30 & 32.93 & 37.75 & 35.18 & 32.76 & 18.75 & 18.89 & 17.88 \\
PCS \cite{yue2021prototypical} & 63.47 & 53.21 & 41.68  & 48.83 & 46.21 & 42.10 & 25.12 & 24.65 & 23.80 \\
Ours & \textbf{66.42} & \textbf{56.84} & \textbf{46.72} & \textbf{52.58} & \textbf{50.10} & \textbf{46.11}& \textbf{39.72} & \textbf{38.59} & \textbf{37.63} \\
\hline

\multirow{2}{*}{}
& \multicolumn{3}{c|}{Quickdraw$\rightarrow$Painting} & \multicolumn{3}{c|}{Quickdraw$\rightarrow$Real}& \multicolumn{3}{c}{Real$\rightarrow$Quickdraw}\\
\cline{2-10}
& P@50 & P@100 & P@200 & P@50 & P@100 & P@200& P@50 & P@100 & P@200 \\
\hline
ID \cite{wu2018unsupervised} & 21.12 & 19.81 & 18.48 & 28.27 & 27.46 & 26.32 & 23.45 & 22.79 & 22.01 \\
ProtoNCE \cite{li2020prototypical} & 23.95 & 22.84 &  21.56 & 26.38 & 25.70 & 24.45 & 25.10 & 24.81 & 23.78 \\
CDS \cite{kim2021cds} & 21.37 & 21.44 & 19.46 & 19.28 & 19.14 & 18.67 & 15.36 & 15.57 & 15.82 \\
PCS \cite{yue2021prototypical} & 24.03 & 23.24 & 22.13  & 34.82 & 33.92 & 31.73 & 28.98 & 28.85 & 28.16 \\
Ours & \textbf{33.45} & \textbf{33.81} & \textbf{34.29} & \textbf{42.79} & \textbf{42.75} & \textbf{42.70}& \textbf{41.90} & \textbf{42.10} & \textbf{41.59} \\
\hline

\multirow{6}{*}{Average}
& \multicolumn{3}{c|}{ID \cite{wu2018unsupervised}} & \multicolumn{3}{c|}{ProtoNCE \cite{li2020prototypical}}& \multicolumn{3}{c}{CDS \cite{kim2021cds}} \\
\cline{2-10}
& P@50 & P@100 & P@200 & P@50 & P@100 & P@200& P@50 & P@100 & P@200 \\
\cline{2-10}
& 37.07 & 33.34 & 28.72 & 37.21 & 32.59 & 27.85 & 36.89 & 31.84 & 26.69 \\
\cline{2-10}
 & \multicolumn{3}{c|}{PCS \cite{yue2021prototypical}} & \multicolumn{3}{c|}{Ours}& \multicolumn{3}{c}{Improvement} \\
\cline{2-10}
& P@50 & P@100 & P@200 & P@50 & P@100 & P@200& P@50 & P@100 & P@200 \\
\cline{2-10}
& 41.23 & 36.87 & 31.74 & \textbf{47.09} & \textbf{43.47} & \textbf{39.09} & \textbf{+5.86} & \textbf{+6.59} & \textbf{+7.35} \\
\hline
\end{tabular}
}
\label{domainnet-retrieval}
\end{center}
\end{table}

\vspace{1mm}
\keypoint{Results.} From the retrieval results in Table \ref{office-retrieval}, we make the following observations: 1) ID \cite{wu2018unsupervised} and ProtoNCE \cite{li2020prototypical} are designed for single domain feature representation learning. The domain gap hurts their performance when applied on cross-domain data. 2) Among all the baseline methods, PCS \cite{yue2021prototypical} performs the best. 3) Our proposed method outperforms nearly all the baselines for all pairs in all three evaluation metrics. This shows the effectiveness of our proposed in-domain cluster-wise contrastive learning and the $\operatorname{DD}$ loss. 4) The retrieval accuracy is related to the domain gap. The retrieval accuracy is higher when the domain gap of the pair is smaller.
In the Office-home dataset, Product and Real are the two with the smallest domain gap. Thus, it can be seen that the retrieval accuracy of both Product $\rightarrow$ Real and Real $\rightarrow$ Product are higher than the others. 5) The accuracy for P@1 is always higher than P@5 and P@15, which means it is more likely to retrieve an image from a wrong category when the number of retrieved image becomes larger. 

\begin{figure}[!t]
\centering
\includegraphics[width=1\columnwidth]{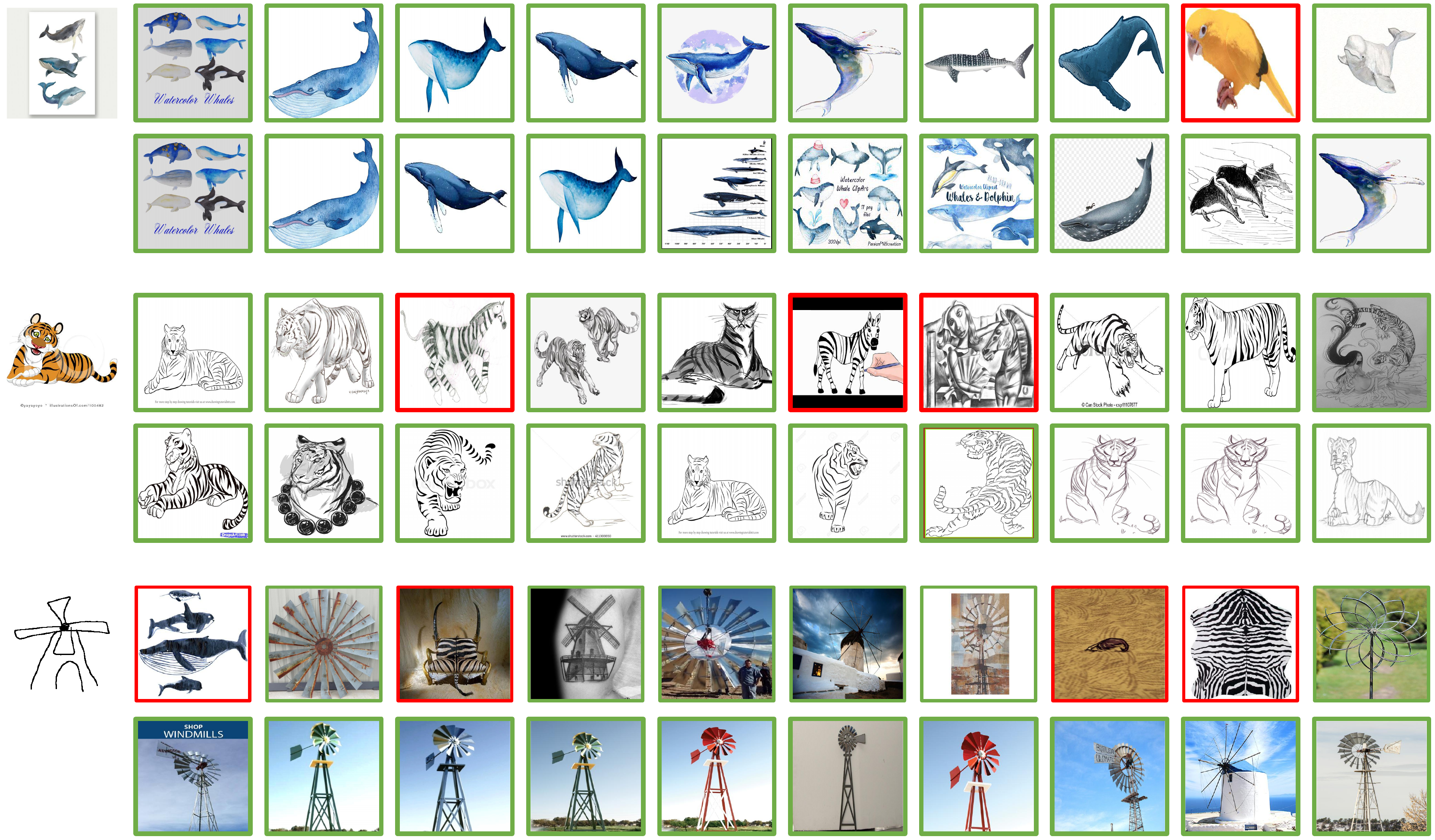}
\label{illustration}
\caption{Top 10 retrieval results on DomainNet. Row 1, 3, 5: Retrieval results of the best baseline method - PCS \cite{yue2021prototypical}; Row 2, 4, 6: Retrieval results of our framework. The green and red boxes indicate correct and incorrect retrievals, respectively.}
\end{figure}

\vspace{2mm}
\superkeypoint{ii) The DomainNet Dataset}

\vspace{1mm}
\keypoint{Settings.} 
We report the results for 6 pairs of domains from the six domains in DomainNet dataset: Clipart-Sketch, Infograph-Real, Infograph-Sketch, Painting-Clipart, Painting-Quickdraw, Quickdraw-Real. We ensure that every domain appears twice for comprehensive evaluations in all six domains.

\vspace{1mm}
\keypoint{Results.} The retrieval performance in Table \ref{domainnet-retrieval} shows that: 1) Similar to the results from Office-Home dataset, PCS \cite{li2020prototypical} is the strongest baseline in our unsupervised cross-domain image retrieval task. 2) Our framework achieves the highest retrieval accuracy for almost all the 12 retrieval tasks. 3) The Quickdraw domain in DomainNet only contain some simple strokes, which lead to the largest domain gap. All the other four baseline methods perform poorly on the Painting $\rightarrow$ Quickdraw, Quickdraw $\rightarrow$ Painting, Quickdraw $\rightarrow$ Real and Real $\rightarrow$ Quickdraw retrieval. 4) The improvement brought by our proposed method is most significant on Painting $\rightarrow$ Quickdraw retrieval. Ours is 14.60$\%$ higher in terms of P@50 when compared to the best baseline PCS. 5) Among all 12 retrieval pairs, our method performs the best for Painting $\rightarrow$ Clipart retrieval and achieves 66.42$\%$ for P@50. 

\begin{table}[!t]
\begin{center}
\caption{Ablation Study on our model component. Cross-domain Retrieval Accuracy $(\%)$ on DomainNet.}
\scalebox{0.93}{
\begin{tabular}{r|c|c|c|c|c|c|c|c|c}
\hline
\multirow{2}{*}{Method}
& \multicolumn{3}{c|}{Clipart$\rightarrow$Sketch} & \multicolumn{3}{c|}{Sketch$\rightarrow$Clipart}& \multicolumn{3}{c}{Infograph$\rightarrow$Real}\\
\cline{2-10}
& P@50 & P@100 & P@200 & P@50 & P@100 & P@200& P@50 & P@100 & P@200 \\
\hline
$\mathcal{L}_{IW}$ (v1) & 48.08 & 43.64& 37.17 & 56.09 & 47.38 & 37.24 & 30.51 & 30.22 & 29.68 \\
$\mathcal{L}_{IW}$ + $\mathcal{L}_{CW}$ (v2) & 51.92 & 47.95 & 41.98 & 60.18 & 52.01 & 41.93 & 32.47& 32.00 & 31.25 \\
$\mathcal{L}_{IW}$ + $\mathcal{L}_{CW}$ + $\mathcal{L}_{SE}$ (v3) & 53.17 & 49.33 & 43.55 & 60.66 & 52.85 & 42.44 & 33.80 & 33.19 & 32.27 \\
Our full model  & \textbf{56.31} & \textbf{52.74} & \textbf{47.38} & \textbf{63.07} & \textbf{57.26}& \textbf{48.17}& \textbf{35.52} & \textbf{35.24} & \textbf{34.35} \\
\hline

\multirow{2}{*}{}
& \multicolumn{3}{c|}{Real$\rightarrow$Infograph} & \multicolumn{3}{c|}{Infograph$\rightarrow$Sketch}& \multicolumn{3}{c}{Sketch$\rightarrow$Infograph}\\
\cline{2-10}
& P@50 & P@100 & P@200 & P@50 & P@100 & P@200& P@50 & P@100 & P@200 \\
\hline
$\mathcal{L}_{IW}$ (v1) & 55.85 & 42.36 & 30.85 & 29.90 & 28.16 & 25.18 & 40.84 & 32.93 & 25.95 \\
$\mathcal{L}_{IW}$ + $\mathcal{L}_{CW}$ (v2) & 57.30 & 45.51 & 33.80 & 30.98 & 29.21 & 26.38 & \textbf{43.81} & 35.99 & 27.86 \\
$\mathcal{L}_{IW}$ + $\mathcal{L}_{CW}$ + $\mathcal{L}_{SE}$ (v3)   & 57.08 & 45.88 & 34.17  & 30.99 & 29.25 & 26.34 & 43.58 & 35.83 & 27.55\\
Our full model & \textbf{57.74} & \textbf{46.69} & \textbf{35.47} & \textbf{31.29} & \textbf{29.33} & \textbf{26.54}& 43.66 & \textbf{36.14} & \textbf{28.12} \\
\hline

\multirow{2}{*}{}
& \multicolumn{3}{c|}{Painting$\rightarrow$Clipart} & \multicolumn{3}{c|}{Clipart$\rightarrow$Painting}& \multicolumn{3}{c}{Painting$\rightarrow$Quickdraw}\\
\cline{2-10}
& P@50 & P@100 & P@200 & P@50 & P@100 & P@200& P@50 & P@100 & P@200 \\
\hline
$\mathcal{L}_{IW}$ (v1) & 55.59 &45.19 & 34.46 & 42.12 & 38.96 & 34.50 & 23.10 & 22.52 & 21.47 \\
$\mathcal{L}_{IW}$ + $\mathcal{L}_{CW}$ (v2) & 65.00 & 54.08 & 41.81 & 47.66 & 44.59 & 40.82 & 24.09 & 23.45 & 22.48 \\
$\mathcal{L}_{IW}$ + $\mathcal{L}_{CW}$ + $\mathcal{L}_{SE}$ (v3) & 66.08 & \textbf{57.20} & \textbf{46.88} & \textbf{52.71} & 49.85 & 46.05 & 34.13 & 33.39 & 32.24 \\
Our full model & \textbf{66.42} & 56.84 & 46.72 & 52.58 & \textbf{50.10} & \textbf{46.11}& \textbf{39.72} & \textbf{38.59} & \textbf{37.63} \\
\hline

\multirow{2}{*}{}
& \multicolumn{3}{c|}{Quickdraw$\rightarrow$Painting} & \multicolumn{3}{c|}{Quickdraw$\rightarrow$Real}& \multicolumn{3}{c}{Real$\rightarrow$Quickdraw}\\
\cline{2-10}
& P@50 & P@100 & P@200 & P@50 & P@100 & P@200& P@50 & P@100 & P@200 \\
\hline
$\mathcal{L}_{IW}$ (v1) & 23.11 & 22.22 & 21.20 & 25.62 & 24.98 & 24.05 & 26.83 & 26.52 & 25.53 \\
$\mathcal{L}_{IW}$ + $\mathcal{L}_{CW}$ (v2)  & 24.83 & 23.80 & 22.32 & 32.32 & 31.63 & 30.50 & 28.25 & 27.56 & 26.53 \\
$\mathcal{L}_{IW}$ + $\mathcal{L}_{CW}$ + $\mathcal{L}_{SE}$ (v3)  & 32.86 & 32.35 & 31.45 & 37.12 & 37.11 & 36.63 & 33.19 & 33.11 & 32.54 \\
Our full model & \textbf{33.45} & \textbf{33.81} & \textbf{34.29} & \textbf{42.79} & \textbf{42.75} & \textbf{42.70}& \textbf{41.90} & \textbf{42.10} & \textbf{41.59} \\
\hline

\multicolumn{10}{c}{Average}\\
\hline
 \multicolumn{1}{c|}{$\mathcal{L}_{IW}$} & \multicolumn{3}{c|}{$\mathcal{L}_{IW}$ + $\mathcal{L}_{CW}$}& \multicolumn{3}{c|}{$\mathcal{L}_{IW}$ + $\mathcal{L}_{CW}$ + $\mathcal{L}_{SE}$}&
 \multicolumn{3}{c}{Our full model}\\
 \hline
 
 \multirow{2}{*}{ \begin{tabular}{c|c|c}
P@50 & P@100 & P@200\\
 38.14 & 33.99 & 29.42
 \end{tabular}}
  & P@50 & P@100 & P@200& P@50 & P@100 & P@200 &P@50 & P@100 & P@200 \\
\hline
 & 41.57 & 37.32 & 32.31 & 44.61 & 40.78 & 36.01 &
\textbf{47.09} & \textbf{43.47} & \textbf{39.09}
\\
\hline
\end{tabular}
}
\label{abla-retrieval}
\end{center}
\end{table}

\subsection{Ablation Study}
 The results in Table \ref{abla-retrieval} show the efficacy of different components in our framework. From the table, we can see: 1) Compared to using the instance-wise contrastive learning loss $\mathcal{L}_{IW}$ (v1), our proposed cluster-wise contrastive loss (v2) indeed helps to learn a better feature embedding for cross-domain image retrieval. All three evaluation metrics show the effectiveness of $\mathcal{L}_{CW}$. 2) In v3, we add the self-entropy loss for clustering probabilities. The improvement over v2 shows that the entropy minimization for clustering probabilities is beneficial for cross-domain feature representation learning. 3) Our full model, which employs the $\operatorname{DD}$ loss to minimize the discrepancy between domains, provides the best alignment and performs the best compared with all v1, v2 and v3. 4) The efficacy of the $\operatorname{DD}$ loss varies a lot in the experiment with different pairs. Comparing to v3 (without $\operatorname{DD}$ loss), our full model achieves a performance gain of only 0.66$\%$ at P@50 on Real $\rightarrow$ Infograph, but shows higher performance gain of 8.71$\%$ on Real $\rightarrow$ Quickdraw.

\section{Conclusion}
This paper presents a novel representation learning framework for unsupervised cross-domain image retrieval which is a challenging but practically valuable task. To extract class semantic-aware feature for category-level retrieval, we propose a cluster-wise contrastive learning loss that pulls samples of similar semantics closer and pushes different clusters apart. For cross-domain alignment, a novel distance of distance loss is introduced to effectively measure the discrepancy between domains and minimized to align features in both domains. The experiment results on Office-Home and DomainNet dataset consistently illustrate the superiority of our proposed algorithm.

\vspace{5mm}
\keypoint{Acknowledgements.} This research/project is supported by the National Research Foundation, Singapore under its AI Singapore Programme (AISG Award No: AISG2-RP-2021-024), and the Tier 2 grant MOE-T2EP20120-0011 from the Singapore Ministry
of Education.

\clearpage
%
%
\bibliographystyle{splncs04}
\bibliography{egbib}
\end{document}